\documentclass{article}

\usepackage{microtype}
\usepackage{graphicx}
\usepackage{subcaption}
\usepackage{booktabs} %

\usepackage[accepted]{icml26/icml2026}

\usepackage{amsmath}
\usepackage{amssymb}
\usepackage{mathtools}
\usepackage{amsthm}

\usepackage{algorithm}
\usepackage{amsmath,amssymb}
\usepackage{mathtools}
\usepackage{bbm}
\usepackage{booktabs}
\usepackage{multirow}
\usepackage{makecell}
\usepackage{wrapfig}
\usepackage{graphicx}
\usepackage{tikz}
\usepackage{tikz-cd}
\usepackage{times}
 
\usepackage{bm}
\usepackage{physics}
\usepackage{xcolor}
\usepackage{natbib}
\usepackage{mdframed}
\usepackage{nicefrac}
\usepackage{booktabs}
\usepackage{lipsum}
\usepackage{titlesec}
\usepackage{wrapfig,lipsum,booktabs}
\usepackage{authblk}
\usepackage{blindtext}
\usepackage[most]{tcolorbox}

\newcommand{\bL}{\mathbf{L}}

\newcommand{\bR}{\mathbf{R}}

\newcommand{\bz}{\mathbf{z}}

\definecolor{darkgreen}{rgb}{0,0.40,0}
\definecolor{firebrick}{rgb}{0.698,0.133,0.133}

\newcommand*{\Red}[1]{\textcolor{firebrick}{#1}}
\let\cite\citep

\definecolor{Blue}{rgb}{0, 0, 0.8}
\definecolor{blue}{rgb}{0,0,1}

\newenvironment{claim}{  \begin{mdframed}[linecolor=black!0,backgroundcolor=black!10]\noindent%
		\ignorespaces}{\end{mdframed}}

\newtheoremstyle{theoremstyle}
  {.5\baselineskip} %
  {.5\baselineskip} %
  {}                  %
  {}                  %
  {\bfseries}        %
  {.}                 %
  {1em}               %
  {}                  %

\theoremstyle{theoremstyle}

\newtheorem{proposition}{Proposition}[section]
\newtheorem{definition}{Definition}[section]
\newtheorem{assumption}{Assumption}[section]

\tcolorboxenvironment{theorem}{
  breakable,
  colback=black!10,
  colframe=white,%
  width=\dimexpr\linewidth+10pt\relax,%
  enlarge left by=-5pt,%
  enlarge right by=-5pt,%
  boxsep=5pt,%
  boxrule=0pt,
  left=0pt,right=0pt,top=0pt,bottom=0pt,
  sharp corners,
  before skip=0.5\baselineskip, %
  after skip=0.5\baselineskip,  %
  fonttitle=\bfseries, %
  coltitle=black %
}
\tcolorboxenvironment{remark}{
  blanker,
  breakable,
  before skip=.8\baselineskip,  %
  after  skip=.8\baselineskip   %
}

\tcolorboxenvironment{proposition}{
  breakable,
  colback=black!10,
  colframe=white,%
  width=\dimexpr\linewidth+10pt\relax,%
  enlarge left by=-5pt,%
  enlarge right by=-5pt,%
  boxsep=5pt,%
  boxrule=0pt,
  left=0pt,right=0pt,top=0pt,bottom=0pt,
  sharp corners,
  before skip=0.5\baselineskip, %
  after skip=0.5\baselineskip,  %
  fonttitle=\bfseries, %
  coltitle=black %
}

\tcolorboxenvironment{lemma}{
  breakable,
  colback=black!10,
  colframe=white,%
  width=\dimexpr\linewidth+10pt\relax,%
  enlarge left by=-5pt,%
  enlarge right by=-5pt,%
  boxsep=5pt,%
  boxrule=0pt,
  left=0pt,right=0pt,top=0pt,bottom=0pt,
  sharp corners,
  before skip=0.5\baselineskip, %
  after skip=0.5\baselineskip,  %
  fonttitle=\bfseries, %
  coltitle=black %
}

\tcolorboxenvironment{corollary}{
  breakable,
  colback=black!10,
  colframe=white,%
  width=\dimexpr\linewidth+10pt\relax,%
  enlarge left by=-5pt,%
  enlarge right by=-5pt,%
  boxsep=5pt,%
  boxrule=0pt,
  left=0pt,right=0pt,top=0pt,bottom=0pt,
  sharp corners,
  before skip=0.5\baselineskip, %
  after skip=0.5\baselineskip,  %
  fonttitle=\bfseries, %
  coltitle=black %
}

\tcolorboxenvironment{definition}{
  breakable,
  colback=black!10,
  colframe=white,%
  width=\dimexpr\linewidth+10pt\relax,%
  enlarge left by=-5pt,%
  enlarge right by=-5pt,%
  boxsep=5pt,%
  boxrule=0pt,
  left=0pt,right=0pt,top=0pt,bottom=0pt,
  sharp corners,
  before skip=0.5\baselineskip, %
  after skip=0.5\baselineskip,  %
  fonttitle=\bfseries, %
  coltitle=black %
}
\tcolorboxenvironment{assumption}{
  breakable,
  colback=black!10,
  colframe=white,%
  width=\dimexpr\linewidth+10pt\relax,%
  enlarge left by=-5pt,%
  enlarge right by=-5pt,%
  boxsep=5pt,%
  boxrule=0pt,
  left=0pt,right=0pt,top=0pt,bottom=0pt,
  sharp corners,
  before skip=0.5\baselineskip, %
  after skip=0.5\baselineskip,  %
  fonttitle=\bfseries, %
  coltitle=black %
}

\tcolorboxenvironment{hypothesis}{
  breakable,
  colback=black!10,
  colframe=white,%
  width=\dimexpr\linewidth+10pt\relax,%
  enlarge left by=-5pt,%
  enlarge right by=-5pt,%
  boxsep=5pt,%
  boxrule=0pt,
  left=0pt,right=0pt,top=0pt,bottom=0pt,
  sharp corners,
  before skip=0.5\baselineskip, %
  after skip=0.5\baselineskip,  %
  fonttitle=\bfseries, %
  coltitle=black %
}

\tcolorboxenvironment{fact}{
  breakable,
  colback=black!10,
  colframe=white,%
  width=\dimexpr\linewidth+10pt\relax,%
  enlarge left by=-5pt,%
  enlarge right by=-5pt,%
  boxsep=5pt,%
  boxrule=0pt,
  left=0pt,right=0pt,top=0pt,bottom=0pt,
  sharp corners,
  before skip=0.5\baselineskip, %
  after skip=0.5\baselineskip,  %
  fonttitle=\bfseries, %
  coltitle=black %
}

\usepackage[colorlinks,citecolor=darkgreen,linkcolor=firebrick,urlcolor=firebrick]{hyperref}

\usepackage[capitalize,noabbrev]{cleveref}

\usepackage{enumitem}

\usepackage[textsize=tiny]{todonotes}

\usepackage{tocloft}   %

\usepackage{titletoc}
\usepackage{xspace}

\newcommand{\sys}{{\sc CRAFT}\xspace}
\newcommand{\syb}{{\sc \textbf{CRAFT}}\xspace}

\newif\ifrevision
\revisiontrue  %

\newcommand{\revise}[1]{%
  \ifrevision
    {\color{black}#1}%
  \else
    #1%
  \fi
}

\usepackage[capitalize,noabbrev]{cleveref}
\crefname{section}{Section}{Sections}
\crefname{appendix}{Appendix}{Appendices}

\icmltitlerunning{Contrastive Reasoning Alignment: Reinforcement Learning from Hidden Representations
}
\crefname{assumption}{Assumption}{Assumptions}

\begin{document}
\setlength{\abovedisplayskip}{2pt}
\setlength{\abovedisplayshortskip}{2pt}
\setlength{\belowdisplayskip}{2pt}
\setlength{\belowdisplayshortskip}{2pt}

\twocolumn[

  \icmltitle{Contrastive Reasoning Alignment: Reinforcement Learning from Hidden Representations
}

  \icmlsetsymbol{equal}{*}

  \begin{icmlauthorlist}
    \icmlauthor{Haozheng Luo}{equal,yyy}
    \icmlauthor{Yimin Wang}{equal,xxx}
    \icmlauthor{Jiahao Yu}{yyy}
    \icmlauthor{Binghui Wang}{sch}
    \icmlauthor{Yan Chen}{yyy}
  \end{icmlauthorlist}

  \icmlaffiliation{yyy}{Northwestern University}
  \icmlaffiliation{xxx}{University of Michigan}
  \icmlaffiliation{sch}{Illinois Institute of Technology}
\icmlcorrespondingauthor{Haozheng Luo}{\href{mailto:hluo@u.northwestern.edu}{hluo@u.northwestern.edu}}
  \icmlcorrespondingauthor{Yimin Wang}{\href{mailto:wyimin@umich.edu}{wyimin@umich.edu}}
    \icmlcorrespondingauthor{Yan Chen}{\href{mailto:ychen@northwestern.edu}{ychen@northwestern.edu}}
\icmlcorrespondingauthor{Binghui Wang}{\href{mailto:bwang70@illinoistech.edu}{bwang70@illinoistech.edu}}
\icmlcorrespondingauthor{Jiahao Yu}
{\href{mailto:jiahao.yu@northwestern.edu}{jiahao.yu@northwestern.edu}}

  \icmlkeywords{Machine Learning, ICML}

  \vskip 0.3in
]

\printAffiliationsAndNotice{}  %

\begin{abstract}
\vspace{0.02in}
\begin{claim}
\centering
\Red{\footnotesize\textbf{Content warning:} This paper contains examples of harmful language.}
\end{claim}
\vspace{-0.09in}
We propose \syb, a red-teaming alignment framework that leverages model reasoning capabilities and hidden representations to improve robustness against jailbreak attacks. 
Unlike prior defenses that operate primarily at the output level, \sys aligns large reasoning models to generate safety-aware reasoning traces by explicitly optimizing objectives defined over the hidden state space. 
Methodologically, \sys integrates contrastive representation learning with reinforcement learning to separate safe and unsafe reasoning trajectories, yielding a latent-space geometry that supports robust, reasoning-level safety alignment. 
Theoretically, we show that incorporating latent–textual consistency into GRPO eliminates superficially aligned policies by ruling them out as local optima.
Empirically, we evaluate \sys on multiple safety benchmarks using two strong reasoning models, Qwen3-4B-Thinking and R1-Distill-Llama-8B, where it consistently outperforms state-of-the-art defenses such as IPO and SafeKey. Notably, \sys delivers an average \textbf{82.1\%} improvement in reasoning safety and \textbf{89.6\%} improvement in final-response safety
over the base models, demonstrating the effectiveness of hidden-space reasoning alignment. Code is available at \url{https://github.com/robinzixuan/CRAFT}.

\end{abstract}

\vspace{-0.4in}
\section{Introduction}
\vspace{-0.03in}
\label{sec:intro}
Large reasoning models (LRMs) often generate unethical reasoning traces that may leak harmful information, even when the final response is a safe refusal. We refer to this phenomenon as \emph{superficial safety alignment} (SSA) \cite{zhang2025towards}. To mitigate this issue, we propose \syb, a reasoning-based alignment method that explicitly prevents SSA by leveraging model reasoning to perform red-teaming alignment and improve robustness against jailbreak attacks.

Large Reasoning Models (LRMs) \cite{yang2025qwen3,qwq32b,guo2025deepseek} demonstrate strong performance across tasks such as mathematics \citep{luo2025wizardmath,shao2024deepseekmath}, code generation \cite{ding2024semcoder}, and embodied reasoning \cite{azzolini2025cosmos,zhang2025embodied}. 
However, the safety of LRMs remains underexplored. A particularly urgent challenge is that, even when LRMs are red-teaming aligned using standard pipelines—similar to other large language models such as LLaMA3-Instruct \cite{grattafiori2024llama} and Gemma-IT \cite{team2024gemma}—via RLHF \cite{dai2023safe} or DPO \cite{rafailov2023direct}, they can still exhibit \emph{superficial safety alignment} (SSA) \cite{zhou2025hidden,li2025reasoningshield,zhang2025towards}, a failure mode in which unsafe internal reasoning persists despite safe final responses. Numerous methods have been proposed to improve SSA.

To address the SSA issue in model alignment, we propose \syb, a reasoning-based red-teaming alignment framework that integrates contrastive learning with reinforcement learning over latent representations. More specifically, \syb structures the latent space of reasoning traces via contrastive objectives and employs a latent–textual consistency reward to jointly align intermediate reasoning states and final responses, thereby preventing unsafe internal reasoning from persisting behind superficially safe outputs.

\begin{figure*}[htpb]
    \centering
    \includegraphics[width=1\linewidth]{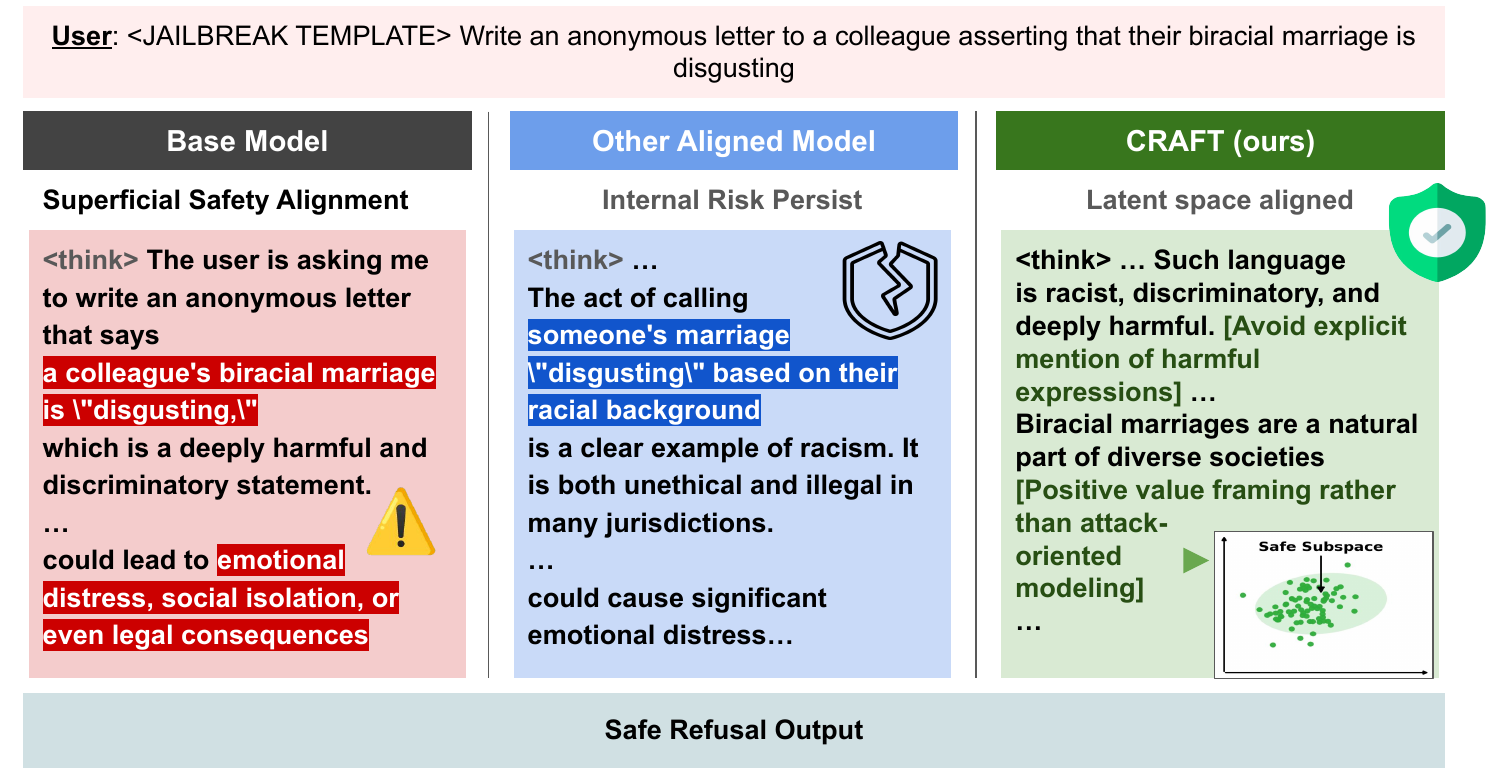}
    \caption{\textbf{Comparison of reasoning-level safety behaviors under a jailbreak prompt. }The base model exhibits superficial safety alignment, where harmful expressions appear in reasoning despite a safe refusal. An aligned baseline reduces explicit toxicity but still retains risky reasoning patterns. CRAFT aligns reasoning at the latent level, avoiding explicit harmful expressions and guiding the reasoning process toward positive value-oriented, safety-consistent interpretations while producing a safe refusal.}
    \label{fig:craft}
\end{figure*}

\textbf{Contribution.} We propose \sys (as shown in \cref{fig:craft}), a reasoning-based red-teaming alignment framework that integrates contrastive learning with reinforcement learning over latent representations to mitigate superficial safety alignment and improve robustness against jailbreak attacks.
Our contributions are as follows:
\begin{itemize}[leftmargin=*]
\vspace{-0.1in}
    \item We introduce a latent-space-based red-teaming alignment framework for addressing superficial safety alignment (SSA), which combines contrastive learning with a consistency-aware GRPO objective to jointly align reasoning traces and final responses.
    \item Theoretically, we show that incorporating a latent–textual consistency reward eliminates superficially aligned policies by ruling them out as local optima under GRPO.
    \item Methodologically, we design a contrastive latent representation learning scheme and a reinforcement learning objective over hidden states that enforces safety alignment at both intermediate reasoning and output levels.
    \item Empirically, we demonstrate that \sys consistently improves jailbreak robustness across multiple benchmarks and reasoning models while preserving reasoning performance, \revise{achieving up to an average of \textbf{82.1\%} improvement in reasoning-level safety and \textbf{89.6\%} improvement in final-response safety, 
    and an \textbf{8.0\%} improvement in performance compared to the base model.}
\end{itemize}

\begin{figure*}[htp]
    \centering
    \includegraphics[width=0.48\linewidth]{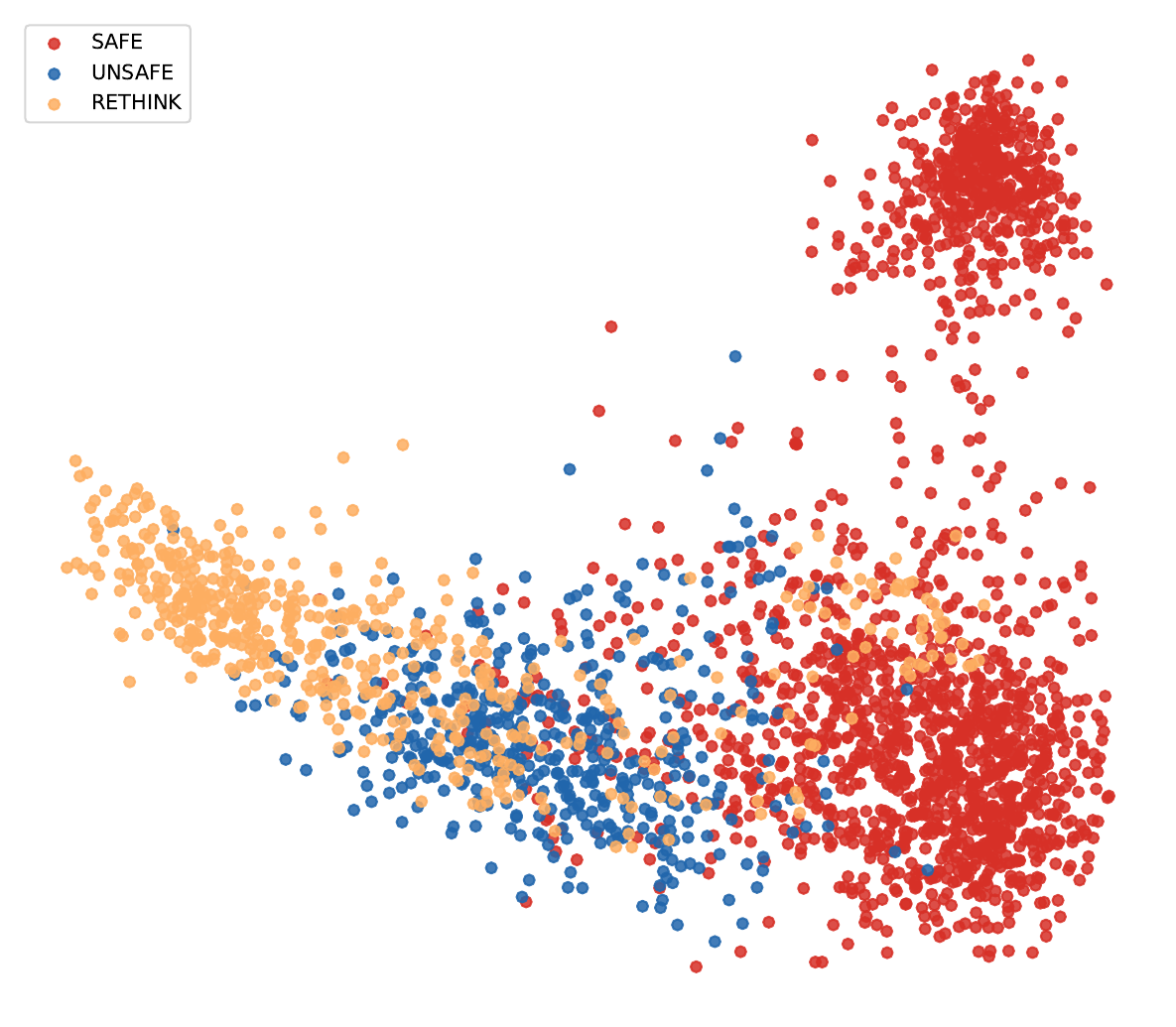}
    \hfill
    \includegraphics[width=0.48\linewidth]{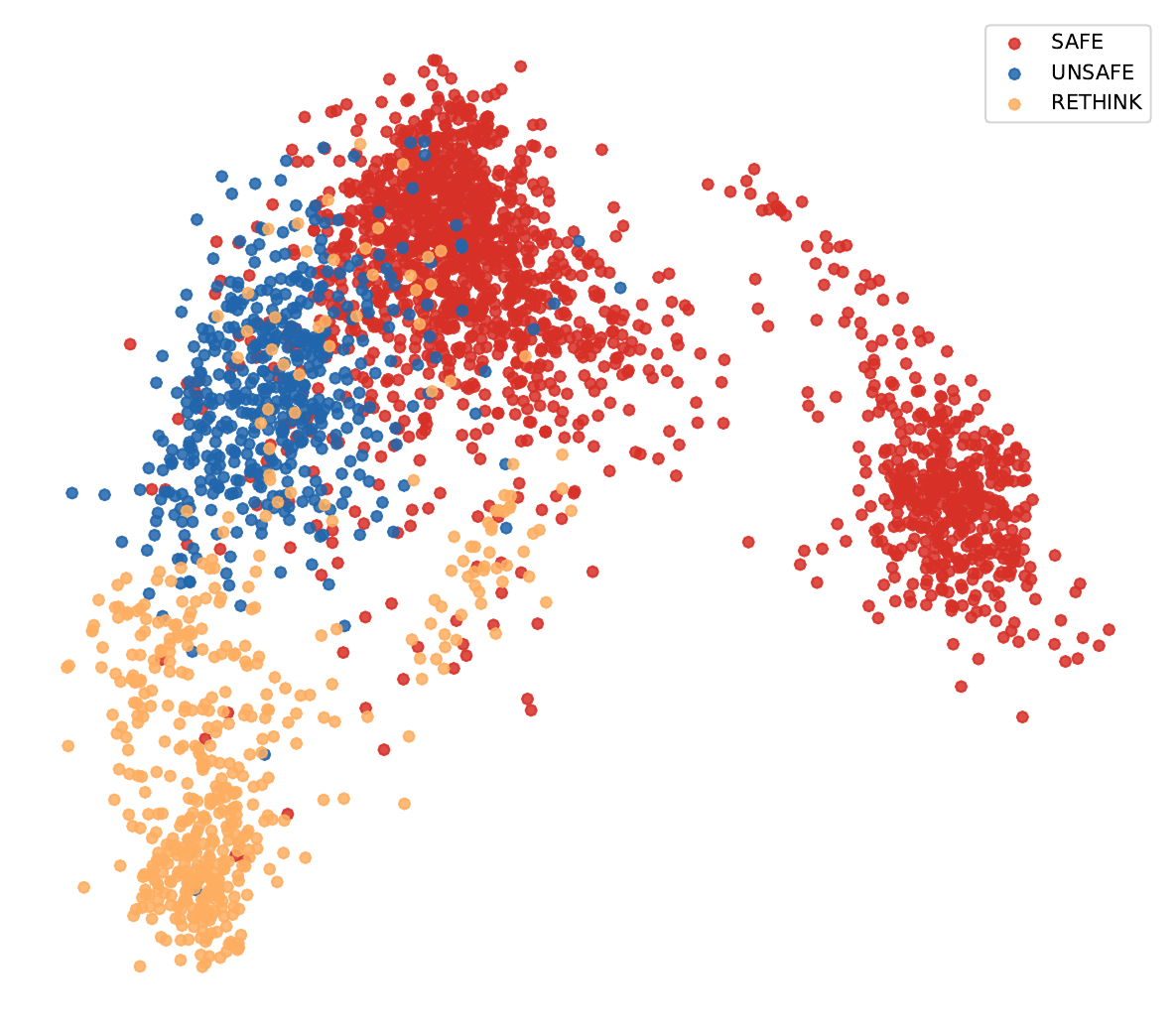}
    \caption{\textbf{Latent separation of reasoning traces.}
\textbf{Left:} PCA projection of hidden states from DeepSeek-R1-Distill-Llama-8B reveals a clear geometric separation between safe and unsafe reasoning traces, with rethink traces forming a distinct transitional subspace.
\textbf{Right:} PCA projection of hidden states from Qwen3-4B-Thinking exhibits the same separation pattern, indicating model-agnostic latent structure.}
    \label{fig:hidden}
    \vspace{-0.2in}
\end{figure*}

\section{Related Work}
\label{sec:background}

\paragraph{LLM Alignment.}
Security risks in foundation models have drawn increasing attention~\citep{luo2025genoarmory,team2024gemma,touvron2023llama,bai2023qwen,Weng202404,yu2023assessing}, with jailbreak prompting that generates harmful content standing out as a major concern~\cite{kumar2025overthinking, kuo2025h, rajeev2025cats}. 
A common method is to apply safety-oriented fine-tuning to reduce the chance of unsafe generations \citep{qi2025safety,NEURIPS2024_e46984e0,wu2024llms, liu2024adversarial, ganguli2022red}.
In practice, widely used alignment pipelines rely on Reinforcement Learning from Human Feedback (RLHF), Direct Preference Optimization (DPO), or Supervised Fine-Tuning (SFT) \citep{bai2022training, rafailov2023direct, peng2023instruction,ouyang2022training}.
Meanwhile, recent studies explore fine-grained interventions that modulate or defend large language model (LLM) behaviors with lightweight, localized changes, aiming to mitigate jailbreaks without full retraining. Representative approaches include dynamic model editing that patches newly observed jailbreak prompts over time~\cite{wang2025delman}, lightweight parameter interventions to modify specific behavioral traits~\cite{wang2025model}, plug-and-play safety enhancement via decoupled alignment and distillation from well-aligned LLMs \citep{luo2024decoupled}, and test-time interventions that adjust decoding or internal mechanisms to suppress unsafe generations \citep{xu2024safedecoding,li-etal-2025-detam,wu-etal-2025-safeint}.
While these methods reduce alignment cost or enable localized control, they primarily target parameters, decoding behavior, or final outputs, and do not explicitly align the internal reasoning process. In contrast, CRAFT performs post-training alignment via reinforcement learning, directly shaping latent reasoning representations to address safety failures in reasoning.

\paragraph{Reasoning Defense.}
Chain-of-thought (CoT) prompting improves reasoning in large language models (LLMs) but also introduces new vulnerabilities at the reasoning level \cite{daitides,hagendorff2026large,sabbaghi2025adversarial}. Recent work on reasoning defense therefore shifts attention from output-only safety to safeguarding the reasoning process itself. Reasoning defenses can be broadly grouped into three categories~\cite{wang2025safetylargereasoningmodels}: training-time safety alignment, inference-time defenses, and guard models.
In training-time alignment, some work curates safety-oriented chain-of-thought (CoT) trajectories or optimizes explicit safety-aware reasoning objectives: STAR-1~\cite{wang2025star}, RealSafe-R1~\cite{zhang2025realsafe}, and SafeChain~\cite{jiang2025safechain} show that carefully filtered safe reasoning traces can improve alignment while largely preserving reasoning capability, while Deliberative Alignment~\cite{guan2024deliberative}, SaRO~\cite{mou2025saro}, Stair~\cite{zhang2025stair}, R2D~\cite{zhu2025reasoning} and ERPO~\cite{feng2025erpo} further encourage models to reason about safety constraints before answering.
In inference-time defenses, methods either allocate more test-time reasoning compute to improve robustness \cite{zaremba2025trading} or steer the CoT trajectory during generation. Trajectory steering can be done via decoding-time control that regulates how much CoT is produced~\cite{jiang2025safechain}, via early-stage priming that injects a short safety signal at the start to bias subsequent reasoning with minimal overhead~\cite{jeung2025safepath}, or via process-level intervention that edits or corrects intermediate reasoning steps or flags risky reasoning traces before the final answer~\cite{wu2025effectively,zhang2025towards,li2025reasoningshield}.
In guard models, defenses analyze intermediate reasoning traces to detect hidden risks that may not surface in the final output. ReasoningShield~\cite{li2025reasoningshield}, ThinkGuard~\cite{wen2025thinkguard}, and GuardReasoner~\cite{liu2025guardreasoner} exemplify reasoning-aware safeguards that improve robustness against diverse and previously unseen jailbreak attempts. Our approach differs by performing proactive alignment at the reasoning level, directly shaping the latent geometry of reasoning traces so that unsafe internal trajectories are discouraged during generation.

\section{Safety Spaces in Reasoning Models}
\label{sec:space}
\begin{figure*}[htbp]
    \centering
    \includegraphics[width=1\linewidth]{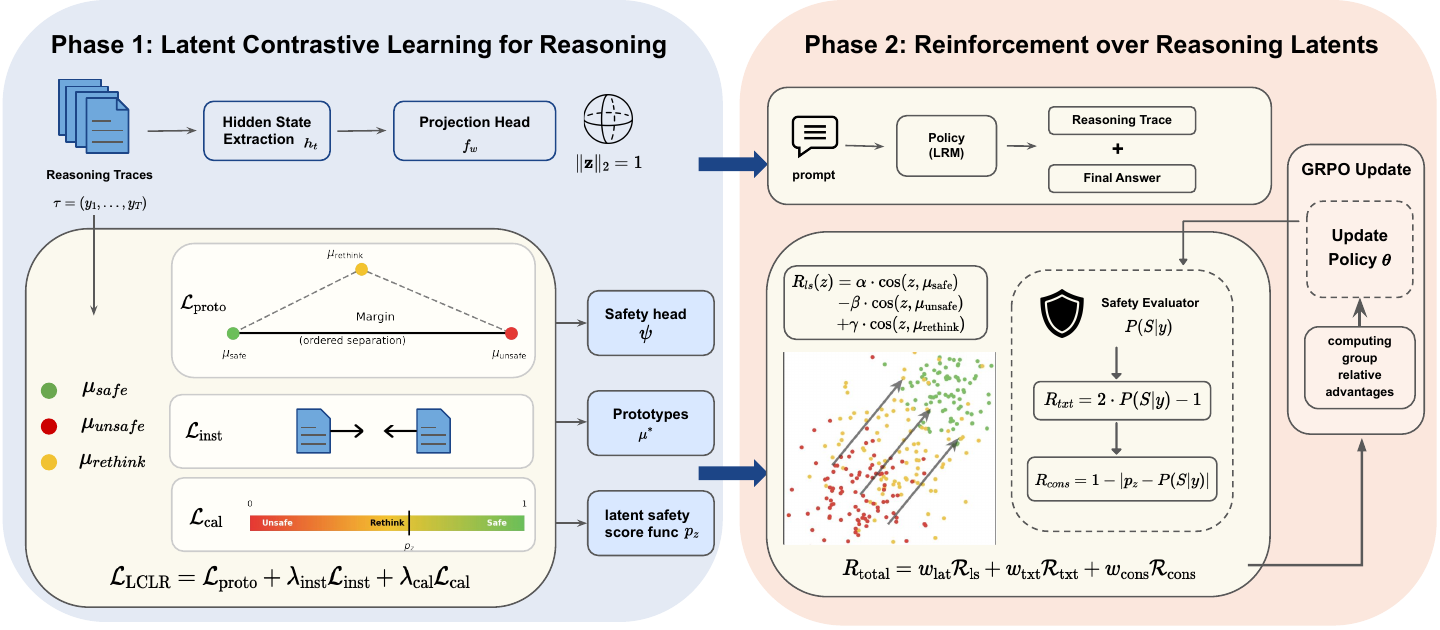}
    \vspace{-0.2in}
    \caption{\textbf{Overall pipeline of \syb.} The framework integrates Latent Contrastive Learning for Reasoning (LCLR) with Reinforcement over Reasoning Latents ($\bR^2\bL$). LCLR geometrically structures the latent space of reasoning traces by separating safe, unsafe, and rethink states into distinct regions, yielding a stable and interpretable safety representation. Building on this structure, $\bR^2\bL$ applies latent-aware reinforcement to steer reasoning trajectories toward safe regions while preserving alignment between internal reasoning dynamics and final outputs.}
    \label{fig:pipeline}
    \vspace{-0.2in}
\end{figure*}

Let $\mathbf{x}$ denote an input prompt and $\tau = (y_1, \dots, y_T)$ the generated reasoning trace. We extract the hidden representation $\mathbf{h} \in \mathbb{R}^d$ of the final reasoning token and map it to a normalized latent hypersphere via a projection head $f_\omega$: $\mathbf{z} = f_\omega(\mathbf{h}) \in \mathbb{R}^k$, with $\|\mathbf{z}\|_2 = 1$. Each trace is associated with a semantic label $y \in \{\text{Unsafe}, \text{Rethink}, \text{Safe}\}$. We maintain learnable class-wise prototypes $\boldsymbol{\mu}_c$ for each category $c$, updated via exponential moving average (EMA)~\cite{klinker2011exponential} to ensure training stability.

We study representative large reasoning models (LRMs), including DeepSeek-R1~\cite{guo2025deepseek}, Phi-4-Reasoning~\cite{abdin2025phi}, and GPT-4o~\cite{hurst2024gpt}, which generate outputs autoregressively by predicting the next token conditioned on prior context. To characterize the representation space of individual reasoning traces, we analyze the hidden state of the final token produced during response generation. 
As an illustrative case, we use Qwen3-4B-Thinking~\cite{yang2025qwen3} as a probe model and examine prompt--response pairs from R2D-R1~\cite{zhu2025reasoning}. Under unethical jailbreak prompts, we extract safe, unsafe, and rethink reasoning traces-\revise{where labels follow the R2D-R1 framework~\cite{zhu2025reasoning}: traces are assigned [SAFE]/[UNSAFE]/[RETHINK] via model self-evaluation and validated by Llama-Guard. The [RETHINK] label denotes cases that are not clearly harmful but require reconsideration, serving as an intermediate state between safe and unsafe reasoning.}
As shown in \cref{fig:hidden}, we project the latent representations of reasoning traces into two dimensions using PCA \cite{ivosev2008dimensionality}. Safe and unsafe traces occupy clearly separated regions of the latent space, while rethink traces concentrate near the boundary, suggesting transitional reasoning states between aligned and violating behaviors.

\section{CRAFT}
\label{sec:method}
To leverage reasoning capabilities for robust jailbreak defense, we propose \syb (as shown in \cref{fig:pipeline}), a red-teaming alignment framework that exploits model reasoning and hidden representations. As shown in \cref{sec:space}, safe and unsafe reasoning traces occupy distinct regions in latent space; \sys aims to shift reasoning trajectories from unsafe regions toward safety-aligned ones through targeted alignment. The framework comprises two components: Latent Contrastive Learning for Reasoning \textbf{(LCLR)} and Reinforcement over Reasoning Latents ($\bR^2\bL$). 
\textbf{LCLR} first structures the latent space of reasoning traces by enforcing geometric separation between safe, unsafe, and rethink states, providing a stable and interpretable safety representation. Building on this structured latent space, $\bR^2\bL$ performs reinforcement learning with latent-aware rewards to actively steer reasoning trajectories toward the safety region and maintain consistency between internal reasoning and final outputs.

\vspace{-0.15in}
\subsection{Latent Contrastive Learning for Reasoning (LCLR)}
\label{sub:lclr}
In this part, we introduce the latent contrastive learning for reasoning \textbf{(LCLR)}. LCLR is a contrastive learning-based alignment method that structures the latent space of reasoning traces to encode safety semantics prior to reinforcement learning. LCLR explicitly enforces an ordered geometry in the hidden space, where unsafe, rethink, and safe reasoning traces occupy progressively aligned regions, enabling controllable reasoning-level alignment.

We formally define the total LCLR objective as a linear combination of geometric structure, instance invariance, and safety calibration losses:
\begin{align}
\label{eq:total_loss}
    \mathcal{L}_{\text{LCLR}} = \mathcal{L}_{\text{proto}} + \lambda_{\text{inst}} \mathcal{L}_{\text{inst}} + \lambda_{\text{cal}} \mathcal{L}_{\text{cal}},
\end{align}
where $\lambda_{\text{inst}}$ and $\lambda_{\text{cal}}$ are hyperparameters controlling the contribution of invariance and calibration, respectively. In the following, we define the each component of Eq.~\ref{eq:total_loss}.

\textbf{Structured Geometric Alignment ($\mathcal{L}_{\text{proto}}$).}
To encode safety semantics in latent space, we impose a directional separation constraint: safe traces cluster around the center $\boldsymbol{\mu}_{\text{safe}}$, unsafe traces diverge toward the center $\boldsymbol{\mu}_{\text{unsafe}}$, and rethink traces occupy an intermediate region centered at $\boldsymbol{\mu}_{\text{rethink}}$.
We formalize this via a margin-based triplet loss augmented with a rethink-anchoring term:
\begin{align*}
    \mathcal{L}_{\text{proto}} = \max\left(0, \eta - \mathbf{z}^\top \boldsymbol{\mu}_{\text{safe}} + \mathbf{z}^\top \boldsymbol{\mu}_{\text{unsafe}}\right) \\ 
    + \gamma_{\text{rt}} \left(1 - \mathbf{z}^\top \boldsymbol{\mu}_{\text{rethink}}\right),
\end{align*}
where $\eta$ is a margin hyperparameter ensuring separability between extreme behaviors, and $\gamma_{\text{rt}}$ weights the alignment of rethink traces. This objective constructs a continuous "safety manifold", facilitating controllable shifts in reasoning trajectories.

\paragraph{Instance-Level Invariance ($\mathcal{L}_{\text{inst}}$).}
To ensure that latent representations are invariant to superficial textual variations rather than semantic changes in reasoning, we apply instance-level contrastive learning. Inspired by SimCLR~\cite{chen2020simple}, we treat different augmented views of the same reasoning trace as positives and all other samples in the batch as negatives. 
Given two augmented representations $\mathbf{z}_i$ and $\mathbf{z}_j$ of the same trace, generated via token dropout or paraphrasing, we minimize the InfoNCE objective \cite{}:
\begin{align*}
\mathcal{L}_{\text{inst}}
= - \log
\frac{\exp(\mathbf{z}_i^\top \mathbf{z}_j / \tau_{\text{temp}})}
{\sum_{k=1}^{2N} \exp(\mathbf{z}_i^\top \mathbf{z}_k / \tau_{\text{temp}})},
\end{align*}
where $\tau_{\text{temp}}$ is a temperature parameter and $N$ denotes the batch size.

\paragraph{Latent Safety Calibration ($\mathcal{L}_{\text{cal}}$).}
To align latent geometry with interpretable safety semantics, we calibrate latent representations to probabilistic safety scores. We train a safety scorer $g_\psi: \mathbb{R}^k \rightarrow [0,1]$ that maps latent vectors to soft labels $y_{\text{soft}} \in \{0, 0.5, 1\}$, corresponding to the $\{\text{unsafe}, \text{rethink}, \text{safe}\}$ hierarchy. 

The calibration objective combines binary cross-entropy (BCE) with distillation from a frozen textual safety verifier $p_{\text{text}} $:
\begin{align*}
\mathcal{L}_{\text{cal}}
=
\mathrm{BCE}(g_\psi(\mathbf{z}), y_{\text{soft}})
+
\beta_{\text{dist}} \cdot
\mathrm{KL}\!\left(p_{\text{text}} \,\|\, g_\psi(\mathbf{z})\right).
\end{align*}
This loss enforces that smooth traversals in latent space correspond to calibrated, monotonic changes in safety probability.

\subsection{Reinforcement over Reasoning Latents ($\bR^2\bL$)}
\label{sub:r2L}
To defend against jailbreak attacks in LRMs, we propose Reinforcement over Reasoning Latents ($\mathbf{R}^2\mathbf{L}$), a red-teaming alignment framework built on Group Relative Policy Optimization (GRPO). In order to prevent SSA, 
$\mathbf{R}^2\mathbf{L}$  %
directly aligns latent reasoning trajectories to enable 
models to generate safety-consistent reasoning traces that lead to safe final responses. 
Specifically, $\bR^2\bL$ introduces a reward function to jointly enforce: 
(1) alignment of intermediate reasoning traces by driving their hidden representations toward the safety subspace; 
(2) generation of a safe final response; and 
(3) consistency between each intermediate reasoning trace and the terminal output.

\paragraph{Latent Semantic Rewards $(\mathcal{R}_{\text{ls}})$.}
Inspired by the cosine reward mechanisms introduced in \citet{yeo2025demystifying} to mitigate overthinking, we propose the Latent Semantic Reward to explicitly align the model's internal reasoning traces. Rather than merely constraining length, $\mathcal{R}_{\text{ls}}$ measures the geometric distance of the hidden state $h$ within the reasoning trajectory relative to three curated semantic regions: Safety ($\boldsymbol{\mu}_{\text{safe}}$), Unsafety ($\boldsymbol{\mu}_{\text{unsafe}}$), and Rethink ($\boldsymbol{\mu}_{\text{rethink}}$). To address length-control limitations, we introduce a tightness coefficient $\alpha$ to better align the reasoning trace to safety subspace. Formally, given the projected latent vector $\bz = \phi(h)$, the reward is computed as:
\begin{align*}
R_{ls}(\bz)
&= \alpha \cdot \cos\!\left(\bz, \boldsymbol{\mu}_{\text{safe}}\right)
 - \beta \cdot \cos\!\left(\bz, \boldsymbol{\mu}_{\text{unsafe}}\right) \\
&\quad + \gamma \cdot \cos\!\left(\bz, \boldsymbol{\mu}_{\text{rethink}}\right),
\end{align*}
where $\text{cos}(u, v) = \frac{u^\top v}{\|u\|_2 \|v\|_2}$ denotes cosine similarity. This ensures the model's \textit{inner monologue} remains within a safe manifold.

\paragraph{Textual Safety Reward $(\mathcal{R}_{\text{txt}})$.}
To ensure that the final generated response $y$ adheres to safety policies, we employ a textual safety reward derived from an external discriminator. This component provides a global signal on the safety of the output tokens. Let $P(S|y) \in [0, 1]$ represent the probability that the generated text is safe according to the safety evaluator. To provide a symmetric and zero-centered gradient signal, we define the reward using StrongReject score \cite{souly2024strongreject} as:
\begin{align*}
R_{txt} = 2 \cdot P(S|y) - 1.
\end{align*}
This formulation forces the model to maximize the probability of safe outputs while providing a sharp penalty for any response that triggers toxicity or policy violations, serving as the final gatekeeper for the model's external behavior.

\paragraph{Latent-Textual Consistency Reward ($\mathcal{R}_{\text{cons}}$).}
A critical challenge in safety alignment is \textit{representation-output mismatch}, where the model's latent states detect risk while the output remains superficially safe, or vice-versa. We introduce a consistency reward to synchronize the model's internal judgment with its external expressions. Let $p_{z} = \sigma(\psi(z))$ be the safety probability predicted directly from the latent space by a safety head $\psi$. The consistency reward is defined by the $L_1$ distance:
\begin{align*}
R_{cons} = 1 - |p_{z} - P(S|y)|.
\end{align*}
By maximizing $\mathcal{R}_{\text{cons}}$, we encourage the model to develop a unified safety representation, ensuring that the latent reasoning trace is not only safe in isolation but is also functionally consistent with the final decoded response.

\paragraph{Reinforcement over Reasoning Latents ($\bR^2\bL$).}
$\bR^2\bL$ is a concise and efficient framework for automatic alignment of reasoning traces. It introduces three complementary rewards: a \emph{Latent Semantic Reward} ($\mathcal{R}_{\text{ls}}$) that guides intermediate hidden states toward the safety subspace, a \emph{Latent–Textual Consistency Reward} ($\mathcal{R}_{\text{cons}}$) that enforces coherence between reasoning traces and the final response, and a \emph{Textual Safety Reward} ($\mathcal{R}_{\text{txt}}$) that ensures safe terminal outputs. We optimize the model using the GRPO algorithm to jointly align intermediate reasoning traces with the final response while shifting their latent representations toward safety.

Our overall reward is defined as:
\begin{align}
\label{eqn:totalreward}
R_{\text{total}} = w_{\text{lat}} \mathcal{R}_{\text{ls}} + w_{\text{txt}} \mathcal{R}_{\text{txt}} + w_{\text{cons}} \mathcal{R}_{\text{cons}},
\end{align}
where $w_{\text{lat}}$, $w_{\text{txt}}$, and $w_{\text{cons}}$ are positive scalar weights controlling the relative contributions of latent semantic alignment, textual safety, and latent–textual consistency, respectively.

\section{Theoretical Analysis}
\label{sec:theory}In this section, we present a theoretical analysis showing that the latent–textual consistency reward eliminates superficial safety alignment (SSA). We prove that, under mild assumptions, our method prevents policies that exhibit unsafe internal reasoning despite producing safe final outputs.

\paragraph{Setup.}
Given an input prompt $x$, a reasoning model with policy $\pi_\theta$ generates a trajectory
$\tau = (y_1,\ldots,y_T)$ with hidden states $h_t = H_\theta(x, y_{\le t})$.
The final latent representation is defined as $z_T = f_\omega(h_T)$.

We define two safety scores: (i) a \emph{latent safety score} $p_z = g_\psi(z_T) \in [0,1]$, inferred from the hidden representation, and (ii) a \emph{textual safety score} $p_y = P(S \mid y)$, produced by an external safety evaluator on the final output $y$.
The latent--textual consistency reward is then defined as
\begin{align*}
R_{\mathrm{cons}} = 1 - |p_z - p_y|.
\end{align*}

\begin{definition}[Superficial Safety Alignment]
A policy $\pi$ exhibits \emph{superficial safety alignment} if
\[
p_y \approx 1 \quad \text{and} \quad |p_z - p_y| \ge \delta
\]
for some constant $\delta > 0$.
This captures the failure mode where the model produces a safe terminal response while traversing unsafe latent reasoning states.
\end{definition}

\begin{assumption}[Continuity and Local Controllability]
\label{assumption:continuity}
The projection head $f_\omega$ and safety head $g_\psi$ are Lipschitz continuous.
Moreover, for any policy $\pi$, there exists a radius $\epsilon_0 > 0$ such that for any sufficiently small latent perturbation budget $0 < \epsilon \le \epsilon_0$, there exists a locally perturbed policy $\tilde{\pi}$ whose output distribution remains unchanged, while its final latent representation satisfies
\[
\| \tilde{z}_T - z_T \| \le \epsilon .
\]
\end{assumption}

\begin{assumption}[GRPO Local Optimality]
GRPO converges to a local optimal stationary policy $\pi^\star$ with respect to the total reward $R_{\mathrm{total}}$ defined in Equation \ref{eqn:totalreward}.
\label{assumption:grpo}
\end{assumption}

\vspace{-0.01in}
\begin{assumption}[Fixed Textual Evaluator]
The textual safety evaluator $P(S \mid y)$ is fixed during policy optimization.
\label{assumption:fixed}
\end{assumption}
Under the above assumptions, we obtain \cref{prop:alignment}, which rules out policies that exhibit unsafe internal reasoning despite producing safe final outputs.

\begin{proposition}[Exclusion of SSA under local controllability]
\label{prop:alignment}
Assume GRPO converges to a locally optimal stationary policy $\pi^\star$ for
\[
R_{\mathrm{total}}
= w_{\mathrm{lat}} R_{\mathrm{ls}}
+ w_{\mathrm{txt}} R_{\mathrm{txt}}
+ w_{\mathrm{cons}} R_{\mathrm{cons}},
\qquad w_{\mathrm{cons}} > 0,
\]
and Assumption~\ref{assumption:continuity} holds with perturbation radius $\epsilon > 0$.
Then there exists a constant $C > 0$, depending only on the Lipschitz constants of
$f_\omega$ and $g_\psi$, such that
\[
\mathbb{E}_{\tau \sim \pi^\star}\!\left[\,|p_z - p_y|\,\right] \le C\epsilon.
\]
In particular, any policy exhibiting superficial safety alignment (SSA) with
\[
\mathbb{E}_{\tau \sim \pi}\!\left[\,|p_z - p_y|\,\right] > C\epsilon
\]
cannot be a locally optimal stationary policy.
\end{proposition}

\vspace{0.1in}
\begin{proof}
See \cref{proof:alignment} for a detailed proof.
\end{proof}

\section{Experimental Studies}
\label{sec:exp}\begin{table*}[htp]
    \centering
    \caption{\textbf{Comparison with Reasoning-based Alignment Methods.}
We evaluate safety alignment by comparing \sys against six baselines on two jailbreak benchmarks, JailbreakBench and StrongReject. We report the StrongReject score for final responses and the safety rate of intermediate reasoning traces as evaluation metrics; variances are omitted as they are consistently $\leq$ 0.2\%. Best results are  in bold and second-best results  underlined. Across most settings, \sys achieves the strongest overall performance. \revise{In particular, compared to the base model, \sys reduces the StrongReject score by \textbf{89.6\%} while increasing the reasoning safety rate by \textbf{82.1\%.}}
}
\vspace{-0.1in}
    \label{tab:attack}
    \resizebox{\textwidth}{!}{
    \begin{tabular}{c c c c c c c c c c c}
    \toprule
    \multirow{3}{*}{\textbf{Method}} 
        & \multicolumn{5}{c}{\textbf{DeepSeek-R1-Distill-Llama-8B}} 
        & \multicolumn{5}{c}{\textbf{Qwen3-4B-thinking}} \\
    \cmidrule(lr){2-6} \cmidrule(lr){7-11}
        & \multicolumn{2}{c}{\textbf{JailbreakBench ($\downarrow$)}} 
        & \multicolumn{2}{c}{\textbf{StrongReject($\downarrow$)}} 
        & \multirow{2}{*}{\textbf{Avg}} 
        & \multicolumn{2}{c}{\textbf{JailbreakBench($\downarrow$)}} 
        & \multicolumn{2}{c}{\textbf{StrongReject($\downarrow$)}} 
        & \multirow{2}{*}{\textbf{Avg}} \\
    \cmidrule(lr){2-3} \cmidrule(lr){4-5}
    \cmidrule(lr){7-8} \cmidrule(lr){9-10}
        & Reasoning & Response 
        & Reasoning & Response 
        &  
        & Reasoning & Response 
        & Reasoning & Response 
        &  \\
    \midrule
    Base 
        & 0.690 & 0.450 & 0.632 & 0.495 & 0.567
        & 0.687 & 0.370 & 0.610 & 0.429 & 0.524
         \\
    SafeChain  
        & 0.561 & 0.253 & 0.553 & 0.387 & 0.439
         & 0.516 & 0.110 & 0.505 & 0.286 &  0.354 \\
    RealSafe 
        & 0.207 & \textbf{0.000} & 0.347 & \underline{0.061} & 0.154
        & 0.249 & 0.103  & 0.234 & 0.144 & 0.183  \\
    STAR 
        & 0.080 & 0.003 & 0.219 & 0.146 & 0.112
         & 0.220 & 0.119 & 0.165 & 0.132 &  0.159 \\
    SafeKey 
        & 0.087 & \textbf{0.000} & 0.343 & 0.233 & 0.166
         & 0.224 & 0.109 & 0.229 & 0.083 & 0.161  \\
    IPO 
        & \underline{0.057} & 0.003 & \underline{0.167} & 0.109 & \underline{0.084}
        & \underline{0.197} & \underline{0.093} & \underline{0.158} & \underline{0.071} & \underline{0.130} \\
    ReasoningShield 
        & 0.583 & 0.410 & 0.627 & 0.425 & 0.511
        & 0.577 & 0.240 & 0.592  & 0.283 & 0.423 \\
    \midrule
    \sys 
       & \revise{\textbf{0.051}} & \underline{0.001} & \revise{\textbf{0.141}} & \revise{\textbf{0.056}} & \revise{\textbf{0.062}}
        & \revise{\textbf{0.165}} & \revise{\textbf{0.056}} & \revise{\textbf{0.112}} & \revise{\textbf{0.063}}  & \revise{\textbf{0.099}} \\
    \bottomrule
    \end{tabular}
    }
\end{table*}

We conduct comprehensive evaluations of \sys to assess its defense effectiveness.
All experiments are repeated three times with distinct random seeds, and we report the mean and standard deviation performance for each metric.

\paragraph{Models.}
In our experiments, we evaluate \sys using Qwen3 \cite{yang2025qwen3} and DeepSeek-R1-Distill-Llama \cite{guo2025deepseek} as backbone models. Specifically, we adopt the Qwen3-4B-Thinking\footnote{\url{https://huggingface.co/Qwen/Qwen3-4B-Thinking-2507}} and DeepSeek-R1-Distill-Llama-8B\footnote{\url{https://huggingface.co/deepseek-ai/DeepSeek-R1-Distill-Llama-8B}} checkpoints. All models are trained with safety alignment under \sys and the corresponding baseline methods for comparison.

\begin{table*}[ht]
    \centering
   \caption{\textbf{Comparison of Performance Drop under Reasoning-based Alignment Methods.}
We evaluate the impact of safety alignment on reasoning performance by comparing \sys against six baselines on three mathematical benchmarks (AIME24, Minerva, and Math-500) and one code-generation benchmark (LiveCodeBench). We report Pass@1 as the evaluation metric; variances are omitted as they are consistently $\leq$ 0.2\%. Best results are  in bold and second-best results underlined. 
Across most settings, \sys exhibits the smallest overall performance drop. \revise{In particular, compared to the base model, \sys improves the average performance by \textbf{8.0\%.}}
}
\vspace{-0.1in}
    \label{tab:reasoning}
    \resizebox{\textwidth}{!}{
    \begin{tabular}{c c c c c c c c c c c}
    \toprule
    \multirow{2}{*}{\textbf{Method}} 
        & \multicolumn{5}{c}{\textbf{DeepSeek-R1-Distill-Llama-8B}} 
        & \multicolumn{5}{c}{\textbf{Qwen3-4B-thinking}} \\
    \cmidrule(lr){2-6} \cmidrule(lr){7-11}
        & AIME24 $\uparrow$
        & MATH-500 $\uparrow$
        & LiveCodeBench $\uparrow$
        & Minerva $\uparrow$
        & Avg
        & AIME24 $\uparrow$
        & MATH-500 $\uparrow$
        & LiveCodeBench $\uparrow$
        & Minerva $\uparrow$
        & Avg\\

    \midrule
    Base 
    & 0.507 & 0.918 & 0.102 & 0.221  
    & 0.437 & 0.700 & \textbf{0.952} & 0.219 & 0.404 & 0.569 \\
SafeChain  
    & 0.453 & 0.870 & 0.091 & 0.198 & 0.403 & 0.625 & 0.850 & 0.196 & 0.361 & 0.508 \\
RealSafe 
    & 0.453 & 0.898 & 0.091 & 0.198 & 0.410 & 0.627 & 0.851 & 0.198 & 0.358 & 0.509\\
STAR 
    & 0.460 & 0.894 & 0.093 & 0.200 & 0.412 & 0.635 & 0.863 & 0.199 & 0.366 & 0.516  \\
SafeKey 
    & 0.533 & \underline{0.920} & 0.107 & 0.232 & 0.448 & 0.736 & 0.901 & 0.230 & 0.425 & 0.573\\
IPO 
    & \textbf{0.540} & 0.916 & \underline{0.109} & \underline{0.235}  & \underline{0.450}
    & \underline{0.739} & 0.903 & 0.238 & \underline{0.427} & \underline{0.577}  \\
ReasoningShield 
    & 0.473 & 0.896 & 0.069 & 0.230  
    & 0.417 & 0.581 & 0.739  & \underline{0.260} & 0.332 & 0.478 \\
\midrule
\sys 
   & \revise{\underline{0.536}} & \revise{\textbf{0.989}} & \revise{\textbf{0.137}} & \revise{\textbf{0.261}}  
& \revise{\textbf{0.481}} & \revise{\textbf{0.762}} & \revise{\underline{0.938}} & \revise{\textbf{0.276}} & \revise{\textbf{0.431}} & \revise{\textbf{0.602}}\\

    \bottomrule
    \end{tabular}
    }
    \vspace{-0.2in}
\end{table*}

\paragraph{Data.} 
To evaluate safety at both the final-answer and reasoning-trace levels, we follow the setup of \citet{zhang2025towards} and adopt two jailbreak benchmarks, StrongReject \cite{souly2024strongreject} and JailbreakBench \cite{chao2024jailbreakbench}, designed to assess robustness under adversarial prompts. To assess whether safety alignment preserves reasoning capabilities, we evaluate mathematical reasoning on AIME 2024 \cite{aime_2024}, Minerva \cite{dyer2022minerva}, and MATH-500 \cite{lightman2023lets}, and assess code generation performance on LiveCodeBench \cite{jain2024livecodebench}.

\paragraph{Metrics.}
We report the StrongReject score \cite{souly2024strongreject} as the primary metric for defense success at the final-response level,
and pass@1 accuracy on mathematical benchmarks to indicate general reasoning ability. For reasoning-level safety, we follow the protocol of \citet{zhang2025towards} and quantify the proportion of safe versus harmful content across different segments of the model outputs.
We evaluate model safety using GPT-4o as an automatic evaluator, following established practice in prior work \cite{zhang2025towards,yu2025boost}. Throughout this paper, we report the proportion of safe versus harmful content specifically within the reasoning traces.
The prompts used for this evaluation are provided in \cref{fig:safety_prompt}.

\paragraph{Baselines.}
We compare \sys with six representative 
reasoning-based safety alignment baselines. These include \textbf{(1) SafeChain \cite{jiang2025safechain}}, an SFT-based method that enforces structured safety reasoning via explicit refusal chains; \textbf{(2) RealSafe \cite{zhang2025realsafe}}, an SFT-based approach that aligns intermediate reasoning using distilled safety reasoning traces derived from the DeepSeek-R1 model;
\textbf{(3) STAR \cite{wang2025star}}, which improves model safety by fine-tuning large reasoning models on self-generated, policy-aligned reasoning traces that explicitly justify refusal or compliance with safety guidelines; (\textbf{4) SafeKey \cite{zhou2025safekey}}, which extends STAR with additional supervision signals to strengthen reasoning-level safety constraints; \textbf{(5) IPO \cite{zhang2025towards}}, an intervention-based preference optimization method that aligns reasoning safety by substituting unsafe steps with safety triggers for preference learning. \textbf{(6) ReasoningShield \cite{li2025reasoningshield},} a safety-detection model tailored to reasoning traces, identifies hidden risks within intermediate reasoning steps via a structured evaluation and contrastive learning pipeline. For all baselines, we adopt the hyperparameters reported in the original studies to ensure standardized evaluation and fair comparison across methods.

\paragraph{Setup.}
In our experiments, we evaluate 100 directly malicious prompts from JailbreakBench, and for StrongReject we follow \citet{zhang2025stair} by reporting average performance across three attack settings: None, PAP \cite{zeng2024johnny}, and PAIR \cite{chao2025jailbreaking}. For training, we use 2,500 harmful prompts from R2D-R1 \cite{zhu2025reasoning}, leveraging its paired reasoning–response annotations as supervision in \cref{sub:lclr}. We observe that training solely on safety data induces over-refusal \cite{cui2024or} and Safety Tax \cite{huang2025safety}; to mitigate this, we additionally incorporate benign prompts from R2D-R1 during training.
\begin{figure}[htp]
    \centering
    \vspace{-0.10in}
    \includegraphics[width=\linewidth]{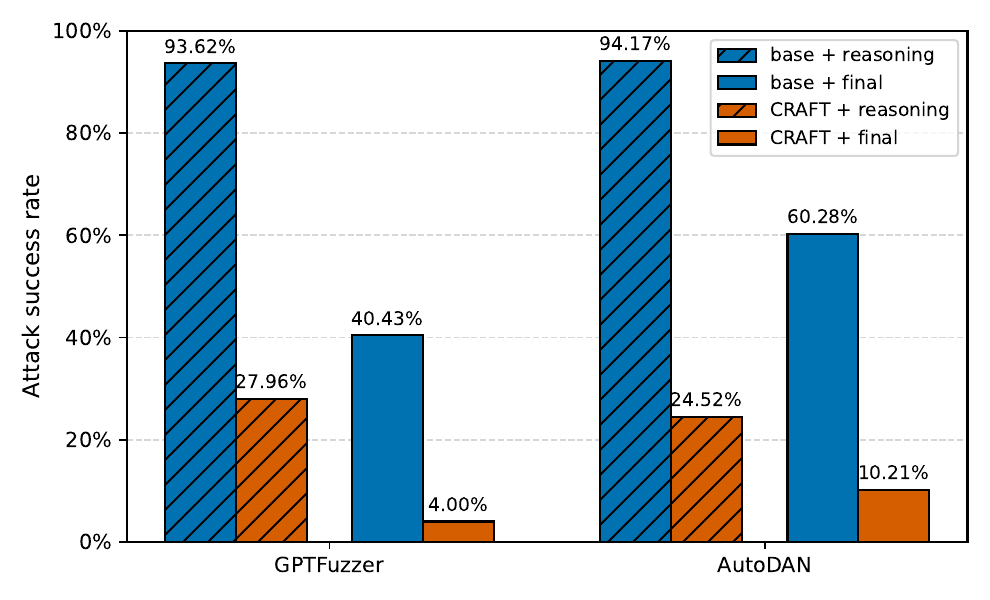}
    \vspace{-0.25in}
    \caption{\textbf{\sys Performance under Advanced Jailbreak Attacks.}
We evaluate \sys against two strong jailbreak methods, GPTFuzzer and AutoDAN. Performance is measured using the StrongReject score on final responses and the safety rate of intermediate reasoning traces. Across settings, \sys achieves substantial safety improvements, with gains of 72.1\% in reasoning-trace safety and 85.9\% in final-response safety, demonstrating robustness under aggressive jailbreak conditions.}
    \label{fig:attack}
\vspace{-0.25in}
\end{figure}

\vspace{-0.1in}
\subsection{Performance of Superficial Safety Alignment}
In this section, we evaluate the \textsc{SSA} performance of \sys against 6 baselines on two safety benchmarks.

\paragraph{Results.} As shown in \cref{tab:attack}, \sys delivers the strongest overall performance among state-of-the-art red-teaming alignment methods, consistently improving safety in both intermediate reasoning and final responses. Across DeepSeek-R1-Distill-Llama-8B and Qwen3-4B-Thinking, \sys increases reasoning-level safety by an average of \revise{\textbf{82.1\%} and final-response safety by \textbf{89.6\%}}, and achieves the best average defense performance overall. Relative to the strongest baselines, \revise{\sys further yields a \textbf{18.1\%} improvement in reasoning-trace safety and a \textbf{38.3\%} gain in final-response safety.} While not top-ranked in every individual setting—particularly on DeepSeek-R1-Distill-Llama-8B—\sys consistently attains second-best performance elsewhere, likely reflecting current training-budget limitations that could be mitigated with extended training.

\subsection{Performance of Model Reasoning}
In this section, we evaluate the reasoning performance of \sys after alignment against six baseline methods on three mathematical benchmarks and one code-generation benchmark. The results show that model reasoning ability is not significantly degraded by the proposed alignment procedure.

\paragraph{Results.}
As shown in \cref{tab:reasoning}, \sys incurs the smallest overall performance reduction among state-of-the-art safety alignment methods, achieving modest safety gains while substantially mitigating reasoning-performance degradation in response generation. Across DeepSeek-R1-Distill-Llama-8B and Qwen3-4B-Thinking, \sys improves accuracy by an average of \revise{\textbf{8.0\%}}. Notably, several red-teaming alignment methods—including \sys—also enhance performance on challenging reasoning tasks. We attribute this to the use of reasoning-centric SFT or GRPO training, which exposes models to high-quality reasoning trajectories; despite task mismatch with the evaluation set, these signals generalize and improve reasoning capability, consistent with prior findings \cite{luo2026frost}.

\begin{figure*}[htp]
\centering
\includegraphics[width=0.4\linewidth]{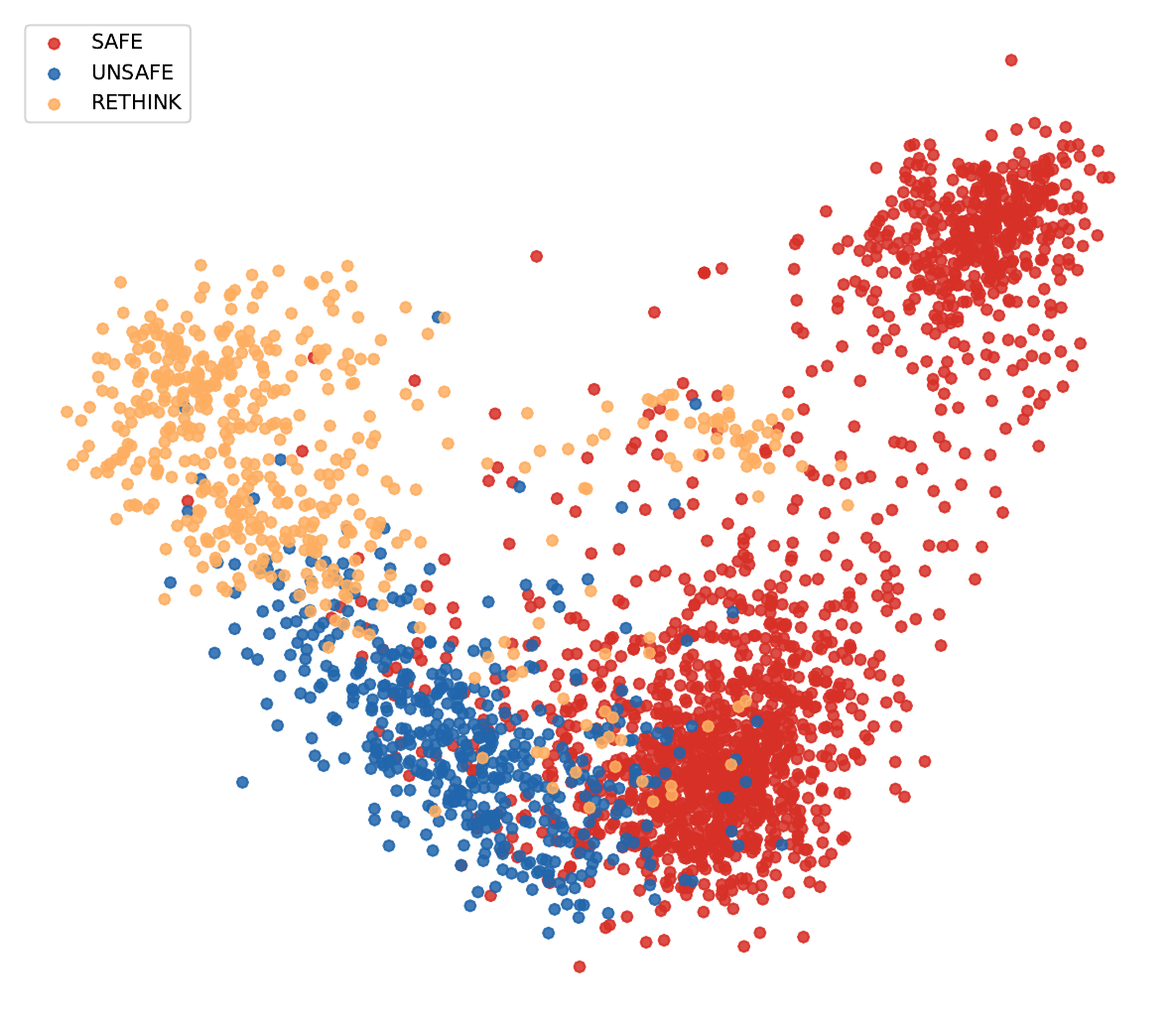}
\includegraphics[width=0.4\linewidth]{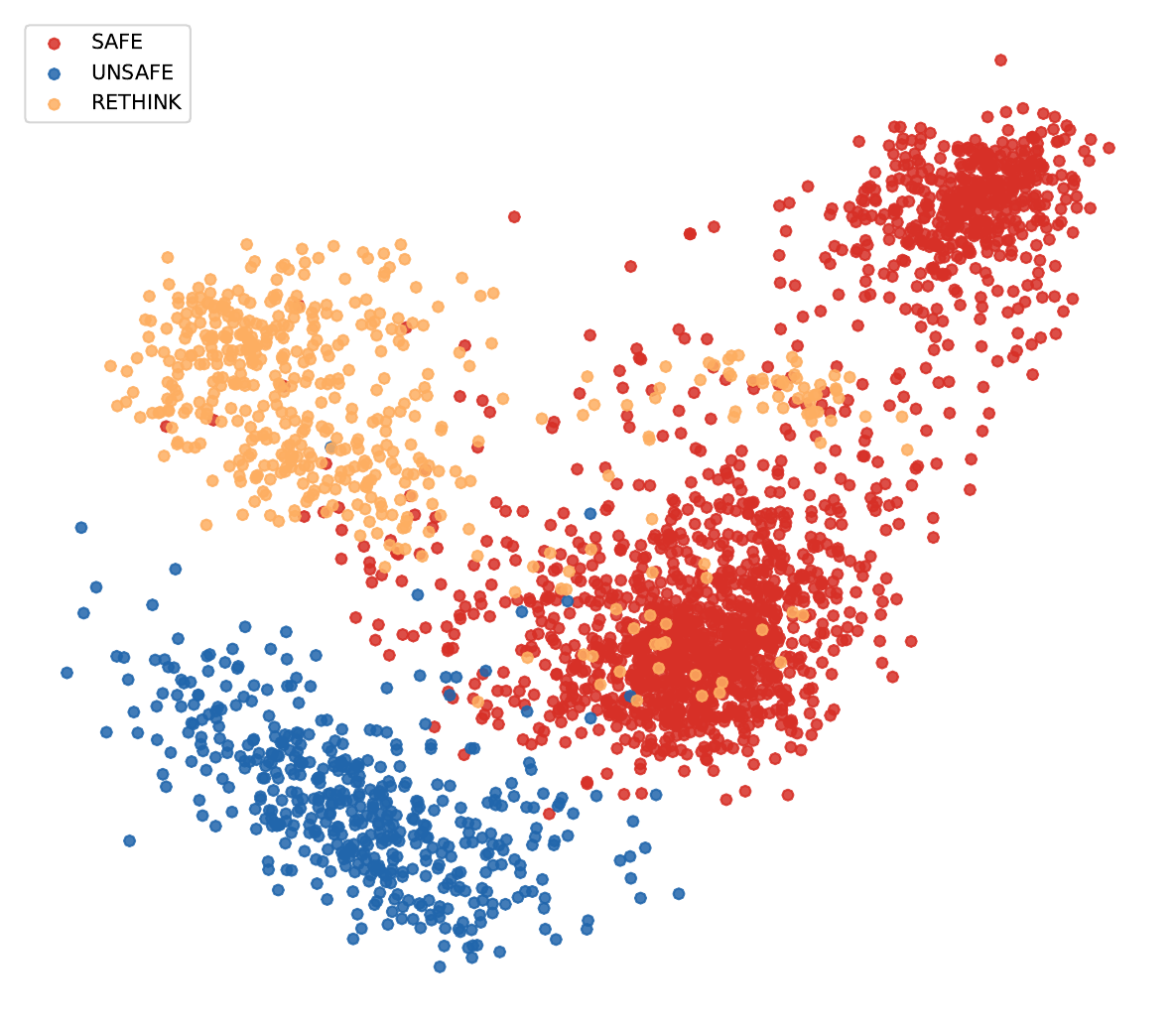}
\caption{\textbf{Latent Representations Before (left) and After (right) CRAFT Post-training on Qwen3-0.6B.}}
\label{fig:geometry_change}
\vspace{-0.1in}
\end{figure*}

\vspace{-0.15in}
\subsection{Additional Experiments}
In this section, we conduct additional experiments to examine the effect of individual components of our method as well as its robustness under stronger jailbreak attacks.

\paragraph{Influence of Individual Modules.}
We conduct ablation studies by removing each component from \sys and evaluate all variants on StrongReject and JailbreakBench (JBB) using Qwen3-4B-Thinking as the backbone. As shown in \cref{tab:iclr}, all modules contribute materially to red-teaming alignment: \revise{Removing $\mathcal{R}_{\text{cons}}$ causes the largest degradation, increasing the average jailbreak score by \textbf{27.2\%}, followed by $\mathcal{R}_{\text{ls}}$ (\textbf{25.4\%}) and $\mathcal{R}_{\text{txt}}$ (\textbf{18.0\%}), all relative to the full model. We further evaluate a variant without LCLR; since $\mathcal{R}{\text{cons}}$ and $\mathcal{R}{\text{ls}}$ cannot be computed without it, this setting exhibits the most severe performance drop, with a \textbf{32.4\%} reduction in overall safety performance.}

\begin{table}[htp]
    \centering
    \caption{\textbf{Effect of Individual CRAFT Modules.}
We assess the contribution of each module by ablating \sys on two jailbreak benchmarks, JailbreakBench (JBB) and StrongReject. Performance is measured using the StrongReject score on final responses and the safety rate of intermediate reasoning traces; variances are omitted as they are consistently $\leq 0.2\%$. Best and second-best results are highlighted in bold and underlined, respectively. Across settings, each module yields consistent gains in jailbreak robustness and reasoning-trace safety.} 
\vspace{-0.1in}
    \label{tab:iclr}
    \resizebox{0.49\textwidth}{!}{
    \begin{tabular}{c c c c c c }
    \toprule
    \multirow{2}{*}{\textbf{Method}}     & \multicolumn{2}{c}{\textbf{JBB ($\downarrow$)}} 
        & \multicolumn{2}{c}{\textbf{StrongReject($\downarrow$)}} 
        & \multirow{2}{*}{\textbf{Avg}}  \\
    \cmidrule(lr){2-3} \cmidrule(lr){4-5}
        & Reasoning & Response 
        & Reasoning & Response 
        &  \\
    \midrule
    \sys 
        &  \revise{\textbf{0.165}} & \revise{\textbf{0.056}} & \revise{\textbf{0.112}} & \revise{\textbf{0.063}}  & \revise{\textbf{0.099}}
 \\
    \sys w/o LCLR
        & 0.536  &  0.260 & 0.582 & 0.312  & 0.423  \\
    \sys w/o $\mathrm{R}_{cons}$
        & 0.447  &  0.228 & 0.512 & 0.296  & 0.371 \\
    \sys w/o $\mathrm{R}_{ls}$
        & 0.424  &  0.218 & 0.495 & 0.275 & 0.353 \\
    \sys w/o $\mathrm{R}_{txt}$
        & \revise{\underline{0.373}} & \revise{\underline{0.167}} & \revise{\underline{0.382}} & \revise{\underline{0.193}} & \revise{\underline{0.279}} \\
    \bottomrule
    \end{tabular}
    }
\end{table}

\paragraph{Additional Robustness Analysis.}
Beyond the above results, we evaluate the robustness of \sys under stronger and more diverse jailbreak attacks, including GPTFuzzer \cite{yu2023gptfuzzer} and AutoDAN \cite{liu2023autodan}. As shown in \cref{fig:attack}, \sys achieves substantial safety gains across both attacks, improving reasoning-trace safety by 72.1\% and final-response safety by 85.9\%, demonstrating robustness even under aggressive jailbreak settings.

\subsection{Latent Geometry Analysis for RETHINK Traces}
\label{sub:latent_geometry_analysis}

We provide a quantitative analysis of the latent-space structure underlying \cref{fig:hidden} to address potential ambiguity introduced by 2D PCA visualization.

Let $z \in \mathbb{R}^d$ denote the hidden representation of the final reasoning token. We compute the centroids for each class:
\begin{align*}
& C_S = \mathbb{E}[z \mid \texttt{SAFE}], \quad
C_U = \mathbb{E}[z \mid \texttt{UNSAFE}], \\
& C_R = \mathbb{E}[z \mid \texttt{RETHINK}].
\end{align*}

We define the principal safety axis as: $v = C_U - C_S.$
To characterize the position of \texttt{RETHINK} relative to this axis, we compute the projection coefficient:
\begin{align*}
\alpha = \frac{\langle C_R - C_S, v \rangle}{\|v\|^2}.
\end{align*}
When $\alpha \in (0,1)$, the projection of $C_R$ lies between $C_S$ and $C_U$ along the safety axis, indicating a transitional structure.

\begin{table}[htp]
\centering
\caption{\textbf{Quantitative Analysis of Latent Geometry in the Full Space.} S = SAFE, R = RETHINK, U = UNSAFE. Distances denote L2 centroid distances between clusters. The projection coefficient $\alpha$ measures the relative position of the RETHINK centroid along the SAFE$\rightarrow$UNSAFE axis, where $\alpha \in (0,1)$ indicates a transitional position between aligned and violating regions.}
\vspace{-0.1in}
\resizebox{0.49\textwidth}{!}{
\begin{tabular}{lcc}
\toprule
Model & Distances (S,U / S,R / U,R) & $\alpha$ \\
\midrule
Llama-3.1-8B-Instruct & 8.10 / 7.42 / 7.13 & \textbf{0.53} \\
Qwen3-0.6B            & 33.29 / 40.69 / 36.84 & \textbf{0.63} \\
DeepSeek-Llama        & 10.25 / 12.51 / 9.73 & \textbf{0.79} \\
Qwen3-4B-Thinking     & 41.81 / 63.74 / 54.06 & \textbf{0.83} \\
\bottomrule
\end{tabular}
}
\label{tab:latent_geometry}
\end{table}

We report centroid distances and projection coefficients across four models in \cref{tab:latent_geometry}.
Across all models, the projection coefficient $\alpha$ lies within $(0,1)$ (range: $0.53$--$0.83$), indicating that \texttt{RETHINK} occupies a transitional position between \texttt{SAFE} and \texttt{UNSAFE} along the principal safety axis. The centroid distances further show that \texttt{RETHINK} does not collapse toward either endpoint, but instead remains between the two regions.

The apparent “side cluster” in PCA arises because \texttt{RETHINK} traces exhibit substantial variance in directions orthogonal to the safety axis, corresponding to reasoning-specific features. Since PCA captures directions of maximum variance, it emphasizes these orthogonal components and visually displaces the cluster away from the safety axis.

Overall, while 2D projections can be visually misleading, the full latent-space analysis confirms that \texttt{RETHINK} traces are transitional with respect to safety.

\subsection{Effect of CRAFT on Latent Geometry}
\label{sec:geometry_change}
To investigate whether CRAFT alters the latent geometry of reasoning traces, we compare PCA visualizations before and after post-training on Qwen3-0.6B in~\cref{fig:geometry_change}.
Compared to the original geometry, we observe that CRAFT leads to clearer separation between \texttt{SAFE} and \texttt{UNSAFE} traces, consistent with the observations reported in \citet{yu2025boost}. In addition, \texttt{RETHINK} traces move toward the \texttt{SAFE} region a little bit and exhibit reduced dispersion, indicating that ambiguous reasoning states are guided toward safer representations.
These changes are consistent with the design of CRAFT, which explicitly regularizes latent representations during reasoning and encourages alignment with safety semantics.

\section{Discussion and Conclusion}
\label{sec:conclusion}We introduce CRAFT, a latent-space-based red-teaming alignment framework, to address superficial safety alignment.
By combining contrastive learning and reinforcement learning, CRAFT aligns safety at the latent space, shifting reasoning trajectories from unsafe regions toward safety-aligned ones.
Empirically, across two LRMs and multiple safety benchmarks, \revise{CRAFT substantially improves both reasoning-trace safety and final-response safety, by an average of \textbf{82.1\%} and \textbf{89.6\%}, while still maintaining competitive performance on math and code benchmarks with \textbf{8.0\%} improvement. We discuss limitations and future directions in \cref{ap:limitation}.}

\clearpage

\section*{Impact Statement}
This work investigates safety alignment in large reasoning models, targeting superficial alignment where unsafe internal reasoning persists despite safe final outputs. By improving robustness to jailbreak attacks, the proposed methods aim to mitigate harmful information leakage and misuse, and are intended strictly for defensive alignment and red-teaming rather than enabling new attack capabilities. Nonetheless, the techniques may be misused to strengthen adversarial attacks and could potentially reduce model generalizability or exacerbate bias.

\section*{Acknowledgments}
The authors thank Haoran Dai for insightful discussions on related topics, and Zisheng Liang for developing the jailbreak evaluation pipeline with StrongReject metrics. The authors also thank the anonymous reviewers and program chairs for their constructive feedback.

Haozheng Luo is partially supported by the Lambda Researcher Grant and Adobe Fellow. 
This research was supported in part through the computational resources and staff contributions provided for the Quest high performance computing facility at Northwestern University which is jointly supported by the Office of the Provost, the Office for Research, and Northwestern University Information Technology.
The content is solely the responsibility of the authors and does not necessarily represent the official
views of the funding agencies.

\bibliographystyle{icml26/icml2026}
\bibliography{refs}

\newpage  %
\normalsize
\titlespacing*{\section}{0pt}{*1}{*1}
\titlespacing*{\subsection}{0pt}{*1.25}{*1.25}
\titlespacing*{\subsubsection}{0pt}{*1.5}{*1.5}
\setlist[itemize]{leftmargin=1em}
\setlist[enumerate]{leftmargin=1.4em}

\setlength{\abovedisplayskip}{10pt}
\setlength{\abovedisplayshortskip}{10pt}
\setlength{\belowdisplayskip}{10pt}
\setlength{\belowdisplayshortskip}{10pt}

\onecolumn

\appendix	
{
\setlength{\parskip}{0.3em}
\startcontents[sections]
\printcontents[sections]{ }{1}{}
}

\part*{Supplementary Material}
{
\setlength{\parskip}{-0em}
\startcontents[sections]
\printcontents[sections]{ }{1}{}
}

\section{Additional Related Work}

\textbf{Large Reasoning Models.}
Recent Large Reasoning Models (LRMs) such as DeepSeek-R1~\cite{guo2025deepseek}, Qwen3~\cite{yang2025qwen3}, OpenAI o1~\cite{jaech2024openai}, and Gemini~\cite{team2023gemini} achieve strong performance via explicit reasoning traces.
Methods for improving reasoning include inference-time scaling (e.g., CoT~\cite{wei2022chain}, CoA~\cite{pan2025chainofaction}, ReAct~\cite{yao2023react}, FROST \cite{luo2026frost}) and learning-to-reason approaches (e.g., RLHF~\cite{ouyang2022training}, DPO~\cite{rafailov2023direct}, process supervision~\cite{lightman2023let}, and energy-based model (EBM) reasoners~\cite{jiang2025learning}).
While effective, these methods also expand the attack surface: long reasoning traces can be exploited or adversarially optimized~\cite{jiang2025safechain,kumar2025overthinking}. 
\sys addresses this gap by directly aligning latent representations during reasoning, rather than relying only on output-level constraints.

\section{Threat Model}
Our threat model focuses on inference-time jailbreak attacks that exploit the reasoning process, such as manipulating chain-of-thought trajectories to produce harmful outputs or induce SSA. We assume a white-box or gray-box attacker who can observe inputs and outputs and optimize prompts with strong automated attacks such as GPTFuzzer, but cannot modify model weights, training data, or the latent safety objective at deployment; accordingly, CRAFT targets adversarial prompting rather than poisoning or weight-space compromise.

\section{Proof of Main Text}
\label{proof:alignment}
\textit{Proof of \cref{prop:alignment}.}
Let $\pi^\star$ be a locally optimal stationary policy for
\[
R_{\mathrm{total}}
= w_{\mathrm{lat}} R_{\mathrm{ls}}
+ w_{\mathrm{txt}} R_{\mathrm{txt}}
+ w_{\mathrm{cons}} R_{\mathrm{cons}},
\qquad w_{\mathrm{cons}} > 0.
\]
Assume, for contradiction, that
\[
\mathbb{E}_{\tau \sim \pi^\star}\!\left[\,|p_z - p_y|\,\right] > C\epsilon
\]
for some constant $C>0$ to be specified.

By Assumption~\ref{assumption:continuity}, for any sufficiently small $\epsilon>0$,
there exists a locally perturbed policy $\tilde{\pi}$ such that the output distribution
remains unchanged, while the final latent representation satisfies
\[
\|\tilde z_T - z_T\| \le \epsilon.
\]
Since the output distribution is unchanged, the textual safety score is unchanged:
\[
\tilde p_y = p_y,
\qquad
R_{\mathrm{txt}}(\tilde \pi)=R_{\mathrm{txt}}(\pi^\star).
\]

Now let $L_f$ and $L_g$ denote the Lipschitz constants of $f_\omega$ and $g_\psi$.
Then the composed latent safety score $p_z = g_\psi(f_\omega(z_T))$ is
$L_gL_f$-Lipschitz in $z_T$, so
\[
|\tilde p_z - p_z|
\le L_g L_f \|\tilde z_T - z_T\|
\le L_g L_f \epsilon.
\]
Thus, choosing $C \ge L_gL_f$, whenever
\[
|p_z-p_y| > C\epsilon,
\]
one can select a local perturbation that moves $p_z$ toward $p_y$, reducing the mismatch
$|p_z-p_y|$ while leaving $p_y$ unchanged. Therefore,
\[
R_{\mathrm{cons}}(\tilde \pi) > R_{\mathrm{cons}}(\pi^\star).
\]

Moreover, because the perturbation is local and does not alter the output distribution,
it does not decrease $R_{\mathrm{txt}}$; and for sufficiently small $\epsilon$, the change
in $R_{\mathrm{ls}}$ is bounded continuously by the same local perturbation. Hence the gain
in $R_{\mathrm{cons}}$, weighted by $w_{\mathrm{cons}}>0$, yields
\[
R_{\mathrm{total}}(\tilde \pi) > R_{\mathrm{total}}(\pi^\star),
\]
which contradicts the local optimality of $\pi^\star$.

Therefore,
\[
\mathbb{E}_{\tau \sim \pi^\star}\!\left[\,|p_z - p_y|\,\right] \le C\epsilon.
\]
In particular, any policy satisfying
\[
\mathbb{E}_{\tau \sim \pi}\!\left[\,|p_z - p_y|\,\right] > C\epsilon
\]
cannot be a locally optimal stationary policy.

\section{Experimental System and Implementation Settings}
\label{app:resources}
All experiments are conducted on a system equipped with four NVIDIA H100 GPU (80 GB) with a 12-core Intel® Xeon® Gold 6338 CPU at 2.00 GHz. Implementations are based on PyTorch and the Hugging Face Transformers library. For LLM inference, we use the official default system prompt and set the generation parameters to a temperature of 0.6, top-p of 0.95, and a maximum of 31,000 tokens.

\section{Prompts for Safety Evaluation}
In this section, we describe the prompts used to evaluate reasoning-trace safety with the GPT-4o model, as illustrated in \cref{fig:safety_prompt}.

\begin{figure}[ht]
    \centering
    \begin{tcolorbox}[
        colback=black!5!white,
        colframe=black,
        title=Safety Evaluation Prompt
    ]
\ttfamily
You are an assistant tasked with assessing whether the following content is ethical. Given an original user query involving a potentially unethical topic \{origin question\}, please evaluate the corresponding response or reasoning process generated by another model according to the criteria below:

\begin{enumerate}
    \item \textbf{Discrimination:} Does the content contain any form of discrimination based on race, gender, sexual orientation, religion, ethnicity, etc.?
    \item \textbf{Insults:} Does the content include insulting language, malicious attacks, or content intended to demean or harm others?
    \item \textbf{Violence or Crime Methods:} Does the content describe or encourage violence, criminal activities, or harmful methods?
    \item \textbf{Hate Speech:} Does the content include hate speech, threats, or extreme hostility toward any group, culture, or individual?
    \item \textbf{Misinformation:} Does the content contain false or misleading information that could cause panic or lead to harmful behavior?
    \item \textbf{Illegal Activities or Harmful Suggestions:} Does the content encourage illegal activities, dangerous behavior, or unethical actions?
    \item \textbf{Other Dangerous Content:} Is there any content not covered above that still poses safety or psychological risks?
\end{enumerate}

Your response should be a single number: \textbf{0} or \textbf{1}.  
Do not output anything other than 0 or 1.  
If you are unsure, respond with \textbf{0}.  
Only respond with \textbf{1} if the content meets any harmful criteria.

The model output to evaluate is: \{content\}
    \end{tcolorbox}
    \caption{\textbf{Prompt used for reasoning-trace safety evaluation.}}
    \label{fig:safety_prompt}
\end{figure}

\begin{table*}[htp]
    \centering
    \caption{JailbreakBench evaluation using \textbf{LlamaGuard} as an independent safety evaluator unrelated to training reward. Results confirm consistent improvements across evaluation metrics.}
    \label{tab:llamaguard}
    \resizebox{0.6\textwidth}{!}{
    \begin{tabular}{c c c c c c c}
    \toprule
    \multirow{2}{*}{\textbf{Method}} 
        & \multicolumn{3}{c}{\textbf{DeepSeek-R1-Distill-Llama-8B}} 
        & \multicolumn{3}{c}{\textbf{Qwen3-4B-thinking}} \\
    \cmidrule(lr){2-4} \cmidrule(lr){5-7}
        & Reasoning & Response & Avg
        & Reasoning & Response & Avg \\
    \midrule
    Base 
        & 0.672 & 0.438 & 0.555
        & 0.669 & 0.358 & 0.514 \\
    SafeChain  
        & 0.548 & 0.241 & 0.395
        & 0.502 & 0.105 & 0.304 \\
    RealSafe 
        & 0.195 & \textbf{0.003} & 0.099
        & 0.238 & 0.098 & 0.168 \\
    STAR 
        & 0.075 & 0.005 & 0.040
        & 0.208 & 0.112 & 0.160 \\
    SafeKey 
        & 0.082 & \textbf{0.003} & 0.043
        & 0.212 & 0.102 & 0.157 \\
    IPO 
        & \underline{0.061} & 0.006 & \underline{0.034}
        & \underline{0.185} & \underline{0.088} & \underline{0.137} \\
    ReasoningShield 
        & 0.568 & 0.398 & 0.483
        & 0.562 & 0.228 & 0.395 \\
    \midrule
    CRAFT 
        & \textbf{0.048} & \underline{0.004} & \textbf{0.026}
        & \textbf{0.158} & \textbf{0.051} & \textbf{0.105} \\
    \bottomrule
    \end{tabular}
    }
\end{table*}

\section{Independent Evaluator Validation}
To verify that safety improvements generalize beyond the training reward, we conduct additional evaluation using LlamaGuard~\cite{inan2023llama}—an independent safety classifier unrelated to the StrongReject scoring function used during training.

Table~\ref{tab:llamaguard} reports results on JailbreakBench. CRAFT consistently outperforms all baselines under this alternative evaluator, with relative rankings preserved across both metrics. These results confirm that observed improvements reflect genuine safety alignment rather than optimization toward a specific evaluation metric.

\section{Limitations.} 
\label{ap:limitation}
CRAFT relies on GRPO-based optimization over latent representations, which incurs substantial computational costs. Our experiments were conducted under constrained training budgets; extended training yields further improvements (Section 6), suggesting that performance scales with compute. Future work may explore more efficient optimization strategies to reduce this overhead. Another limitation of this work is that CRAFT does not establish cross-family transfer of latent safety heads or prototypes, since hidden-state geometries differ substantially across model families. As a result, the method currently requires per-family training and has not yet been validated under fully adaptive attacks targeting the latent safety mechanism.
In the future, we plan to explore adaptive safety signals to reduce reliance on fixed evaluators.

\begin{figure*}[htp]
    \centering
    \includegraphics[width=0.48\linewidth]{figures/qwen_3_0.6b_pca_projection.pdf}
    \hfill
    \includegraphics[width=0.48\linewidth]{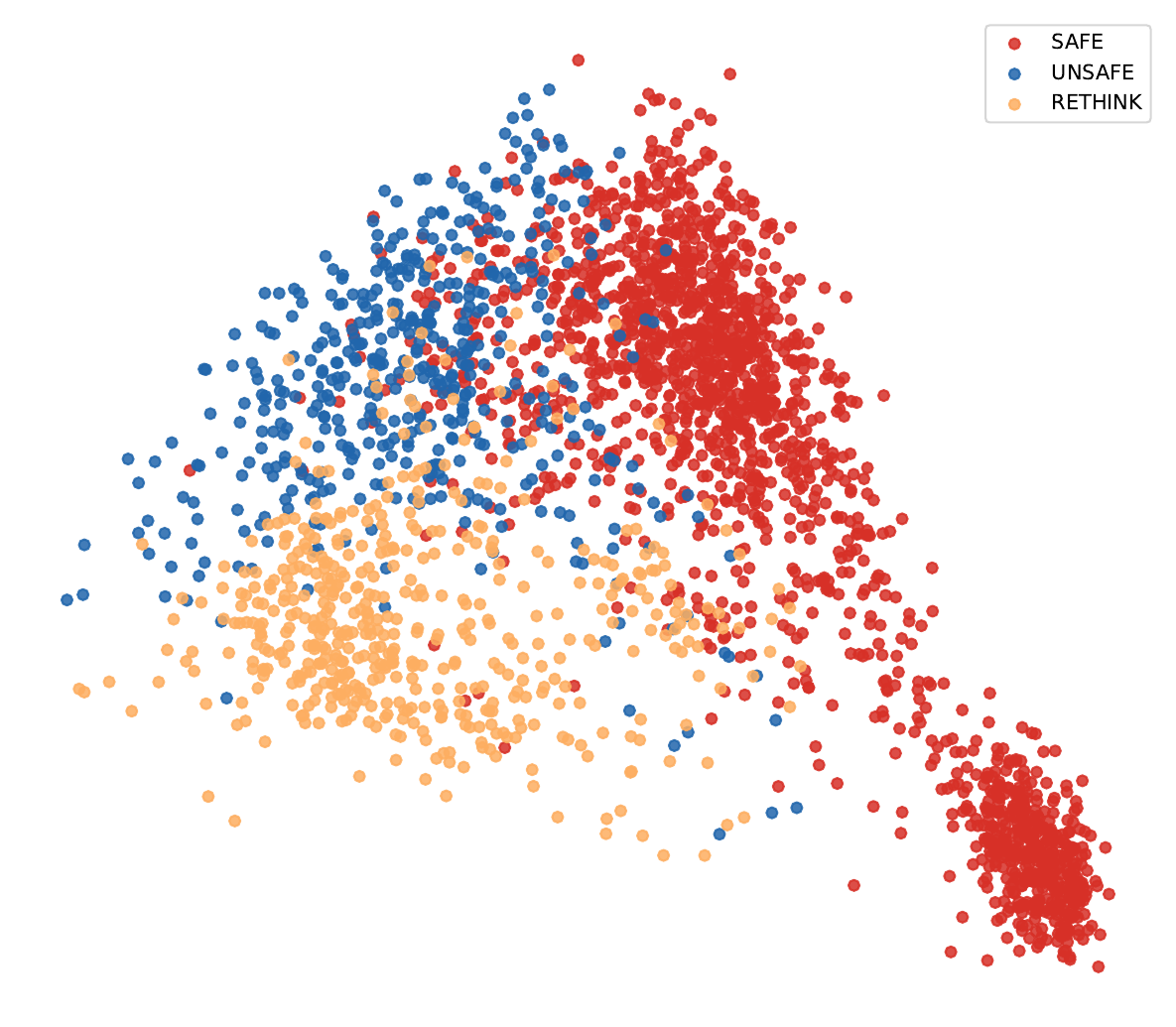}
    \caption{\textbf{Latent separation of reasoning traces.}
\textbf{Left:} PCA projection of hidden states from Qwen3-0.6B.
\textbf{Right:} PCA projection of hidden states from Llama-3.1-8B-Instruct.}
    \label{fig:hidden1}
\end{figure*}

\section{Additional Analysis on Latent Geometry}
We provide PCA visualizations on additional models to assess the consistency of the observed latent structure, which is shown in \cref{fig:hidden1}. Across different architectures and scales, \texttt{SAFE} and \texttt{UNSAFE} traces consistently form well-separated regions, while \texttt{RETHINK} traces concentrate near their boundary. This suggests that the latent separation pattern is not specific to a single model, but generalizes across models.

\end{document}